\documentclass[letterpaper]{article} 
\usepackage{aaai25}  
\usepackage{times}  
\usepackage{helvet}  
\usepackage{courier}  
\usepackage[hyphens]{url}  
\usepackage{graphicx} 
\usepackage{natbib}  
\usepackage{caption} 
\usepackage{algorithm}
\usepackage{algorithmic}

\usepackage{tabularx}
\usepackage{multirow}
\usepackage{xcolor}
\usepackage{colortbl}
\usepackage{amssymb}
\usepackage{amsmath}
\usepackage{pifont}
\usepackage{soul}
\usepackage{booktabs}

\usepackage{newfloat}
\usepackage{listings}
\DeclareCaptionStyle{ruled}{labelfont=normalfont,labelsep=colon,strut=off} 
\floatstyle{ruled}
\newfloat{listing}{tb}{lst}{}
\floatname{listing}{Listing}
\title{Selective Visual Prompting in Vision Mamba}
\author{
    Yifeng Yao, 
    Zichen Liu, 
    Zhenyu Cui, 
    Yuxin Peng, 
    Jiahuan Zhou\thanks{Corresponding Author}
}
\affiliations{
    Wangxuan Institute of Computer Technology, Peking University, Beijing 100871, China\\
    \{yaoyifeng, lzc20180720, cuizhenyu\}@stu.pku.edu.cn, \{pengyuxin, jiahuanzhou\}@pku.edu.cn

}

\usepackage{bibentry}

\begin{document}

\maketitle

\begin{abstract}
Pre-trained Vision Mamba (Vim) models have demonstrated exceptional performance across various computer vision tasks in a computationally efficient manner, attributed to their unique design of selective state space models. To further extend their applicability to diverse downstream vision tasks, Vim models can be adapted using the efficient fine-tuning technique known as visual prompting. However, existing visual prompting methods are predominantly tailored for Vision Transformer (ViT)-based models that leverage global attention, neglecting the distinctive sequential token-wise compression and propagation characteristics of Vim. Specifically, existing prompt tokens prefixed to the sequence are insufficient to effectively activate the input and forget gates across the entire sequence, hindering the extraction and propagation of discriminative information. To address this limitation, we introduce a novel Selective Visual Prompting (SVP) method specifically for the efficient fine-tuning of Vim. To prevent the loss of discriminative information during state space propagation, SVP employs lightweight selective prompters for token-wise prompt generation, ensuring adaptive activation of the update and forget gates within Mamba blocks to promote discriminative information propagation. Moreover, considering that Vim propagates both shared cross-layer information and specific inner-layer information, we further refine SVP with a dual-path structure: Cross-Prompting and Inner-Prompting. Cross-Prompting utilizes shared parameters across layers, while Inner-Prompting employs distinct parameters, promoting the propagation of both shared and specific information, respectively. Extensive experimental results on various large-scale benchmarks demonstrate that our proposed SVP significantly outperforms state-of-the-art methods. Our source code is available at https://github.com/zhoujiahuan1991/AAAI2025-SVP.
\end{abstract}

%

\section{Introduction}

\begin{figure}[!ht]
\centering
\includegraphics[width=0.47\textwidth]{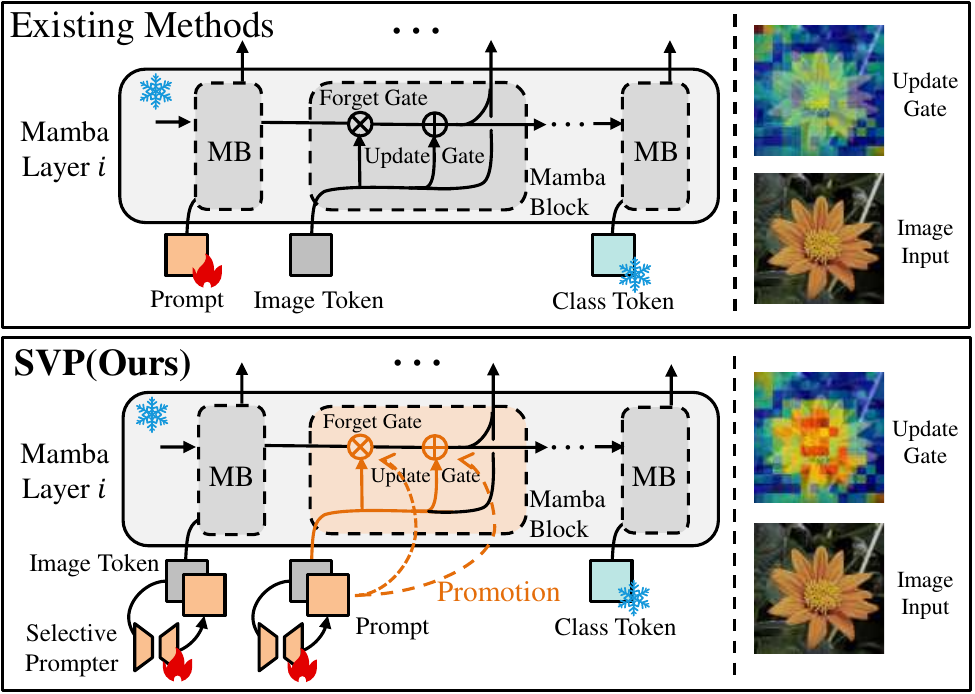} 
\caption{Existing visual prompting methods~\cite{jia2022visual} use prompt sequences prefixed to the image tokens, which hinder discriminative feature propagation in Vim. In contrast, our SVP uses input-dependent selective prompts that better learn the input distribution, activating update and forget gates to enhance object-aware feature propagation. }
\label{fig1}
\end{figure}

Vision Mamba (Vim), a groundbreaking vision backbone featuring an input-dependent modeling mechanism termed selective state space, has demonstrated superior performance compared to established Vision Transformers (ViT) while maintaining computational efficiency~\cite{zhu2024vision}. This advancement positions Vim as a potential next-generation foundation model architecture. However, as models scale up, directly applying pre-trained Vim models to downstream tasks through full fine-tuning leads to significant computational and storage overhead. This issue has spurred the development of Parameter-Efficient Fine-Tuning (PEFT) techniques~\cite{fu2023effectiveness}, which adapt pre-trained models to downstream tasks by fine-tuning only a small subset of parameters or incorporating a few additional ones, thereby substantially reducing storage requirements. 
Among PEFT methods, visual prompting~\cite{han20232vpt} has shown promising performance by integrating a few additional learnable parameters into pre-trained models.

Therefore, visual prompting holds substantial potential for the efficient fine-tuning of Vim, enabling high performance with minimal resource consumption. However, current visual prompting methods are predominantly designed for ViT with global attention mechanisms and fail to account for the unique sequential characteristics of Vim, which processes visual information through token-by-token compression and propagation. As illustrated in Figure~\ref{fig1}, existing approaches employ prompt sequences prefixed to the token sequence, which inadequately adapts the model to the input distribution. This limitation poses a significant challenge, as it impedes the effective updating and propagation of discriminative sequence information when directly applied to Vim. There is a pressing need to explore visual prompting techniques specifically tailored to Vim, making it both a significant research endeavor and a practical necessity.

To address the unique requirements of Vim, we propose a novel Selective Visual Prompting (SVP) method designed to promote discriminative information propagation within the sequence. As depicted in Figure~\ref{fig1}, our approach employs a lightweight prompter that dynamically produces token-level prompts based on varying inputs. These prompts are then integrated into the original image tokens, which adequately adapt the model to the input distribution. This mechanism ensures the selective activation of update and output gates, which are input-dependent parameters in Vim. Consequently, our SVP method enables the model to update and compress relevant discriminative features while propagating them through the network. Simultaneously, it identifies and discards irrelevant distracting information, preventing it from contaminating the compressed state of the sequence. This targeted approach enhances the model's ability to retain and propagate discriminative information.

Furthermore, recognizing that Vim propagates both shared cross-layer information and specific inner-layer information, we introduce a dual-path structure in our selective prompting design, comprising Cross-Prompting and Inner-Prompting mechanisms. This design aims to optimize the propagation of both types of information within Vim. The Cross-Prompting facilitates the transfer of shared information between layers, while Inner-Prompting enhances the flow of layer-specific information within each layer. To account for the varying proportions of these two types of information at different layers, we implement an element-wise scaling factor that dynamically adjusts the emphasis between the two prompts. This dual-path structure ensures a balanced and effective leveraging of both information types, significantly enhancing the model's overall performance. 

To sum up, the main contributions of this work are: (1) To the best of our knowledge, this is the initial exploration of visual prompting within Vim. We introduce a selective prompting approach that leverages Vim's input-dependent characteristics, adaptively activating its input and forget gates to enhance the propagation of discriminative information. (2) In our SVP, a dual-path prompting structure, termed Cross-Inner, is proposed to effectively utilize both cross-layer shared information and inner-layer specific features, ensuring comprehensive and efficient information propagation. (3) Extensive experiments on various datasets demonstrate that our SVP method significantly outperforms existing visual prompting approaches, achieving superior performance with equivalent model size and pre-training datasets.

\section{Related Work}
\subsection{State Space Model}

The state space model (SSM) with linear complexity presented a promising approach for modeling long-range dependencies. Moreover, the Structured State-Space Sequence model~\cite{gu2021efficiently} improved computational efficiency while preserving theoretical strengths through novel parameterization. Expanding on this foundation,  Mamba~\cite{gu2023mamba} and Mamba2~\cite{dao2024transformers} introduced a data-dependent SSM layer with hidden state expansion, forming a language model backbone. Building on its success in sequence data, Vision Mamba~\cite{zhu2024vision} applied pure Mamba layers to vision tasks, utilizing bidirectional scans for comprehensive modeling. Vim's linear complexity and effective performance highlighted its suitability for a large pre-trained model. The previous paradigm for adapting pre-trained models to downstream tasks was full fine-tuning. However, as the scale of Vim models increased, this approach became inefficient, driving the development of Parameter-Efficient Fine-Tuning methods (PEFT)~\cite{fu2023effectiveness}.

\subsection{Parameter-Efficient Fine-Tuning}

    PEFT aimed at reducing learnable parameters while maintaining performance, categorized into partial-based, addition-based, and prompt-based methods~\cite{xin2024parameter}. Partial-based methods trained only select portions of model parameters, e.g., bias terms, attention, or MLP layers~\cite{zaken2021bitfit, kornblith2019better, touvron2022three, basu2024strong}. While these approaches are straightforward and simple to implement, they often lag in performance compared to full finetuning. Addition-based methods, such as Side-Tuning~\cite{sung2022lst} and adapters~\cite{chen2022adaptformer, steitz2024adapters, xin2024vmt, dong2024efficient}, introduce auxiliary components for task-specific learning, yet their custom nature limits generalizability across different architectures.

\begin{figure*}[!t]
\centering
\includegraphics[width=0.95\textwidth]{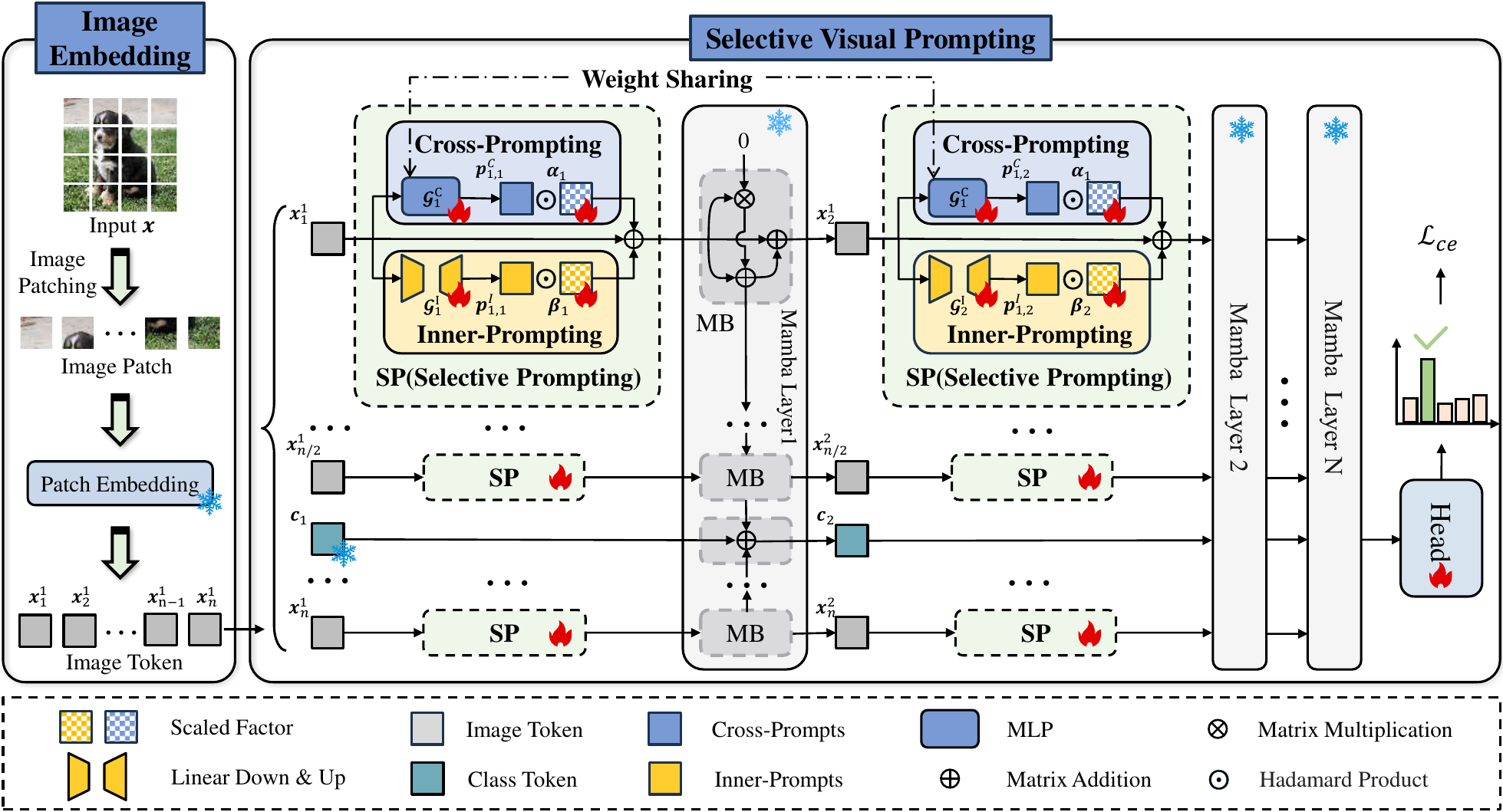} 
\caption{Our SVP employs a dual-path architecture: the Inner-Prompting pathway prompts specific information at each layer, while the Cross-Prompting pathway prompts shared information across layers. Both the Inner-Prompt ${\boldsymbol{p}}^{\scriptscriptstyle I}_i$ and Cross-Prompt ${\boldsymbol{p}}^{\scriptscriptstyle C}_i$ are selectively generated based on the input. They are subsequently coordinated by two element-wise dynamic factors~$\boldsymbol{\alpha}_j, \boldsymbol{\beta}_i$ and then superimposed onto the original input.}
\label{fig2}
\end{figure*}

\subsection{Prompt Learning}

    Visual prompt learning techniques, which operated primarily on the input, offered better generalization and were more compatible with various models. Existing visual prompt learning methods could be categorized into two types. The first type, represented by the VPT series, appended prompt tokens to the image token sequence. E\textsuperscript{2}VPT~\cite{han20232vpt} further refined these prompts by pruning ineffective ones, while more recent approaches like InsVP~\cite{liu2024insvp} learned prompts more relevant to the instance. SPT~\cite{wang2024revisiting} revisited the power of VPT and extended it by self-initializing with downstream token prototypes. However, these methods struggled to effectively capture the distribution across the entire sequence in the Vim sequence model. The second type directly overlayed frame-like prompts onto the original image, as seen in DAM-VP~\cite{huang2023diversity} and AutoVP~\cite{tsao2024autovp}. These methods only applied prompts at the image level and lacked input dependency. As a result, they were not effective at activating the update and forget gates in the deeper layers of Vim. Therefore, directly applying these two types of prompt learning methods to Vim led to suboptimal performance. In contrast, our SVP method selectively activates the update gate across the entire sequence, promoting discriminative information propagation.

\section{The Proposed Method}
\subsection{SSM Preliminaries}

    The SSM-based models, Mamba~\cite{gu2023mamba}, and Vision Mamba (Vim)~\cite{zhu2024vision} are inspired by the continuous system, which maps a one-dimensional function or sequence $x(t) \in \mathbb{R}\mapsto y(t) \in \mathbb{R}$ through a {\rm N}-dimension hidden state $h(t) \in \mathbb{R}^{\rm N}$. The hidden state evolves over time with parameters $\mathbf{A}$, $\mathbf{B}$, and $\mathbf{C}$, following linear ordinary differential equations:


\begin{equation}
    \label{EQ1}    
    \begin{aligned}h'(t)&=\mathbf{A}h(t)+\mathbf{B}x(t),\\y(t)&=\mathbf{C}h(t),\end{aligned}
\end{equation}
where $\mathbf{A} \in \mathbb{R}^{{\rm N}\times {\rm N}}$ is the state matrix, $\mathbf{B} \in \mathbb{R}^{{\rm N}\times 1}$, and $\mathbf{C} \in \mathbb{R}^{1\times {\rm N}}$ are projection parameters.

    To adapt SSM for deep learning, it is discretized using zero-order hold~\cite{pechlivanidou2022zero}. The continuous parameters $\mathbf{A}$, $\mathbf{B}$ are transformed into their discrete counterparts $\overline{\mathbf{A}}\in\mathbb{R}^{{\rm N}\times {\rm N}}$, $\overline{\mathbf{B}}\in\mathbb{R}^{{\rm N}\times 1}$ using a timescale parameter $\Delta \in \mathbb{R}$:

\begin{equation}
    \label{EQ2}
    \begin{aligned}&\overline{\mathbf{A}}=\exp(\Delta\mathbf{A}),\\&\overline{\mathbf{B}}=(\Delta\mathbf{A})^{-1}(\exp(\Delta\mathbf{A})-\mathbf{I})\cdot\Delta\mathbf{B}\approx\Delta\mathbf{B}.\end{aligned}
\end{equation}

    Thus, the discrete SSM can be written as:

\begin{equation}
    \begin{aligned}&{h}_i=\overline{\mathbf{A}}{h}_{i-1}+\overline{\mathbf{B}}{x}_i,\\&{y}_i=\mathbf{C}{h}_i,\end{aligned}
    \label{EQ3}
\end{equation}
 where~$~{h}_{i-1},~{h}_{i}\in\mathbb{R}^{{\rm N}\times 1}, x_i.$

\subsection{Selective Visual Prompting}
    
    To address the challenge of efficient fine-tuning in Vim, we propose a novel approach termed Selective Visual Prompting (SVP). Our method effectively adapts the model to the input distribution by selectively generating prompts at the token level. It adaptively activates the update and forget gates to promote effective information propagation. The overall structure of the proposed method is illustrated in Figure~\ref{fig2}.

    Specifically, the process begins with dividing the input image $\boldsymbol{x} \in \mathbb{R}^{{\rm H}\times {\rm W} \times {\rm C}}$ into equally sized patches, where $({\rm H}, {\rm W})$ represents the size of the image $\boldsymbol{x}$, and ${\rm C}$ is the number of channels. These patches are then embedded into $d$-dimensional latent space as~$\{\boldsymbol{x}_i^p\}_{i=1}^n, \boldsymbol{x}_i\in \mathbb{R}^{1 \times d}$.

    Given Vim's 24-layer hierarchical architecture, each layer in Vim should focus not only on specific inner-layer features but also on shared features with adjacent layers. To facilitate both information propagation, we designed the lightweight Selective Prompting Module with a dual-path structure, incorporating Cross-Prompting and Inner-Prompting.

     \subsubsection{Cross-Prompting.} To capture and propagate shared information across layers to ensure feature consistency, we design a Cross-Prompting Module. In this module, we utilize a fully connected cross-prompts generator~($\mathcal{G}^{\scriptscriptstyle C}$) with shared parameters across layers to generate cross-prompts~$\boldsymbol{p}^{\scriptscriptstyle C}_{i}\in\mathbb{R}^{1\times d}$ through $\boldsymbol{x}_i\in\mathbb{R}^{1\times d}$. The number of layers sharing parameters is a hyperparameter set to 6, 8, or 12. The process is represented by the following formula:

\begin{equation}
   \label{EQ4}
    \boldsymbol{p}^{\scriptscriptstyle C}_i = \mathcal{G}^{\scriptscriptstyle C}(\boldsymbol{x}_i).
\end{equation}

    \subsubsection{Inner-Prompting.} This module focuses on extracting and preserving layer-specific features to enhance the model's discriminative power. To minimize tunable parameters while maintaining performance, we design a lightweight inner-prompts generator~($\mathcal{G}^{\scriptscriptstyle I}$) for each layer of Vim, enabling the generation of distinct inner-prompts~$\boldsymbol{p}^{\scriptscriptstyle I}_i\in\mathbb{R}^{1\times d}$. This generator includes a linear down layer~($L^{down}$)
    , linear up layer~($L^{up}$)
    and a SiLU activation~\cite{elfwing2018sigmoid}. The hidden dimension, which refers to both the reduced dimension in $L^{down}$ and the dimension to be expanded in $L^{up}$, is set to 64.  This process is represented by the following formula:

\begin{equation}
   \label{EQ5}
    \begin{aligned}
    \boldsymbol{p}^{\scriptscriptstyle I}_i = & \mathcal{G}^{\scriptscriptstyle I}(\boldsymbol{x}_i) \\
     =& \text{SiLU}(L^{up}(L^{down}(\boldsymbol{x}_i))).
    \end{aligned}
\end{equation}

    Otherwise, considering the differing importance of layer-specific features and shared information across layers, we designed two element-wise dynamic scaling factors~($\boldsymbol{\alpha}$,~$\boldsymbol{\beta}$) to balance the influence of these prompts on the input distribution. 
    
\begin{equation}
   \label{EQ6}
    \begin{aligned}
    {\boldsymbol{\bar{p}}}_i = & {\boldsymbol{\alpha}}\odot \boldsymbol{p}^{\scriptscriptstyle C}_i + {\boldsymbol{\beta}}\odot\boldsymbol{p}^{\scriptscriptstyle I}_i,
    \end{aligned}
\end{equation}
where~$\boldsymbol{\alpha}\in\mathbb{R}^{1\times d}$ and $\boldsymbol{\beta}\in\mathbb{R}^{1\times d}$ are learnable parameters which are initialized to zero,  $\odot$ denotes Hadamard product, $\bar{\boldsymbol{p}}_i\in\mathbb{R}^{1\times d}$.
As shown in Equation~\ref{EQ7}, the generated prompts are then overlaid onto the original input, effectively activating the update and forget gates in Vim to promote the propagation of shared and layer-specific information. 
    
\begin{equation}
    \label{EQ7}
    \boldsymbol{x}_i^p=\boldsymbol{x}_i+{\boldsymbol{\bar{p}}}_i.
\end{equation}

Subsequently, all prompted image tokens $\{\boldsymbol{x}_i^p\}_{i=1}^n$ along with an additional classification token $\boldsymbol{c}_1 \in  \mathbb{R}^{1 \times d}$ are fed into the $N$ Mamba blocks $\{\mathcal{B}_j\}_{j=1}^N$ to extract features. Similar to the design of VPT-deep, we incorporate our selective prompts at the input of each layer. The output $\boldsymbol{c}_{N+1}$ from the final Mamba block is then passed through a classification head $\mathcal{H}$ to produce the predicted probability distribution $\boldsymbol{y}$.

\subsubsection{Overall Optimization.} As mentioned above, our SVP introduces only a few additional parameters: 
\begin{equation}
    {\boldsymbol{\mathcal{M}}} = \left\{\mathcal{G}^{\scriptscriptstyle C}, \mathcal{G}^{\scriptscriptstyle I}, \boldsymbol{\alpha}, \boldsymbol{\beta}\right\}.
    \label{EQ8}
\end{equation}

Following prior works~\cite{jia2022visual, huang2023diversity, wang2024revisiting}, we keep the pre-trained model's encoder frozen during training, allowing only the classification head~$\mathcal{H}$ and the newly added modules~$\mathcal{M}$ to be trainable. The optimization objective is defined as follows:

\begin{equation}
    \label{EQ9}
    \mathop{\arg\min}\limits_{\boldsymbol{\mathcal{M}}, \mathcal{H}}
        \mathcal{L}_{ce} \left(\boldsymbol{y}, y_{gt} \right),
\end{equation}
where $\mathcal{L}_{ce}$ is the cross-entropy loss, and $y_{gt}$ is the image label.

\begin{figure}[t]
    \centering
    \includegraphics[width=0.25\textwidth]{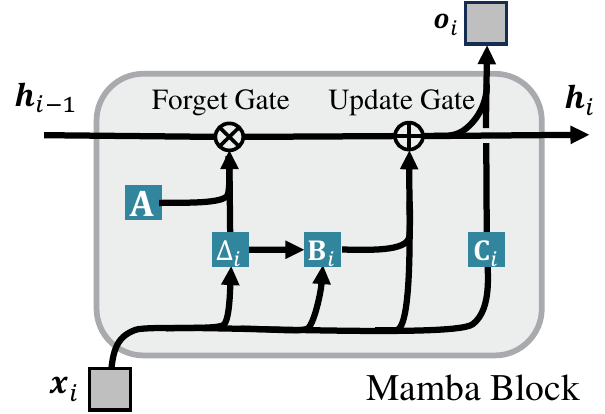} 
    \caption{The internal structure of the Mamba block. The parameters~$\Delta_i$, $\mathbf{B}_i$, and~$\mathbf{C}_i$ are all input-dependent.}
    \label{fig3}
\end{figure}

\subsection{Discussion and Analysis}

    In this section, we discuss how our SVP facilitates the update and forget gates across the entire sequence, thereby promoting effective information propagation.

    As shown in Figure~\ref{fig3}, the Mamba architecture enhances the SSM by introducing the selective state space model. The parameters $\mathbf{B}_i\in\mathbb{R}^{h\times 1}$, $\mathbf{C}_i\in\mathbb{R}^{1\times h}$, and $\Delta_i\in\mathbb{R}^{1\times d}$ are generated from $\boldsymbol{x}_{i}$ via functions $\mathcal{S}_{B}$, $\mathcal{S}_{C}$, and $\mathcal{S}_{\Delta}$, thus becoming input-dependent:

\begin{equation}
    \label{EQ10}
    \begin{aligned}
    \mathbf{B}_i=\mathcal{S}_{B}(\boldsymbol{x}_i), 
    ~\mathbf{C}_i=\mathcal{S}_{C}(\boldsymbol{x}_i), 
    ~\Delta_i=\mathcal{S}_{\Delta}(\boldsymbol{x}_i).
    \end{aligned}
\end{equation}

    Mamba pracatically applies the state transition Equation~\ref{EQ3} independently to each channel of input $\boldsymbol{x}_i$, leading to the following formulations:

\begin{equation}
\label{EQ11}
\begin{aligned}
    \boldsymbol{h}_i&= \mathbf{\widetilde{A}}_i\odot\boldsymbol{h}_{i-1} +\mathbf{B}_i(\Delta_i\odot\boldsymbol{x}_i)\\
    &={\exp}(\mathcal{S}_{\Delta}(\boldsymbol{x}_i)\widetilde{\odot}\mathbf{A})\odot\boldsymbol{h}_{i-1}+\mathcal{S}_{B}(\boldsymbol{x}_i)(\mathcal{S}_{\Delta}(\boldsymbol{x}_i)\odot\boldsymbol{x}_i),
\end{aligned}
\end{equation}
where $\mathbf{A}, \mathbf{\widetilde{A}}_i, \boldsymbol{h}_{i-1}, \boldsymbol{h}_{i}\in\mathbb{R}^{h\times d}$, $\widetilde{\odot}$ denotes extending the first dimension of the preceding matrix, followed by a Hadamard product with the subsequent matrix.

    In our SVP, Vim's parameters~$\mathbf{B}_i$,~$\mathbf{C}_i$~and~$\Delta_i$~are generated through prompted inputs~$\boldsymbol{x}_i^p$~as~$\mathbf{B}_i^p\in\mathbb{R}^{h\times 1}$,~$\mathbf{C}_i^p\in\mathbb{R}^{1\times h}$~and~$\Delta_i^p\in\mathbb{R}^{1\times d}$. Then the state transition equation in Vim can be rewritten as:
    
\begin{equation}
\label{EQ12}
\begin{aligned}
    \boldsymbol{h}_i=& \mathbf{\widetilde{A}}_i^p\odot\boldsymbol{h}_{i-1} +\mathbf{B}_i^p(\Delta_i^p\odot\boldsymbol{x}_i^p)\\
    =&{\exp}(\mathcal{S}_{\Delta}(\boldsymbol{x}_i+\boldsymbol{\bar{p}}_i)\widetilde{\odot}\mathbf{A})\odot\boldsymbol{h}_{i-1}\\
    &+\mathcal{S}_{B}(\boldsymbol{x}_i+\boldsymbol{\bar{p}}_i)(\mathcal{S}_{\Delta}(\boldsymbol{x}_i+\boldsymbol{\bar{p}}_i)\odot\boldsymbol{x}_i)\\
    &+\mathcal{S}_{B}(\boldsymbol{x}_i+\boldsymbol{\bar{p}}_i)(\mathcal{S}_{\Delta}(\boldsymbol{x}_i+\boldsymbol{\bar{p}}_i)\odot\boldsymbol{\bar{p}}_i).
\end{aligned}
\end{equation}

\begin{table*}[h]

\small
    \centering
    \setlength{\tabcolsep}{0.92mm}\renewcommand{\arraystretch}{1.15}\begin{tabularx}{\textwidth}{l|c|c@{\hspace{2pt}}c@{\hspace{1.5pt}}c|cccccccccc|c}
    \toprule
        Methods & Publication & Backbone & Param & Pre-train& Cifar & Cifar10 & DTD & CUB & Dogs & GTSRB & Flowers & SVHN & Birds & Food & \textbf{Average} \\
        \hline
        Full & \textcolor{gray}{\textit{-}} & Vim-S & 25M & 1K & 89.6 & 98.8 & 72.3 & 80.1 & 92.5 & 97.5 & 90.5 & 98.0 & 77.8 & 87.4 & 88.5\\

        \specialrule{0.8pt}{1pt}{1pt}
        \specialrule{0.5pt}{1pt}{1pt}
        
        VP & \textcolor{gray}{\textit{arXiv 2022}} & ViT-B & 85M & 21K & 78.7 & 94.2 & 59.5 & 84.6 & 84.5 & 89.4 & 97.7 & 87.6 & 77.7 & 80.5 & 83.4 \\  

        VPT & \textcolor{gray}{\textit{ECCV 2022}} & ViT-B & 85M & 21K & 78.8 & 96.8 & 65.8 & 88.5 & 90.2 & 90.7 & 99.0 & 78.1 & 84.2 & 83.3 & 85.5 \\
        
        E\textsuperscript{2}VPT & \textcolor{gray}{\textit{ICCV 2023}} & ViT-B & 85M & 21K & 80.4 & 97.1 & 66.8 & 89.1 & 90.5 & 91.0 & 99.1 & 79.2 & 84.6 & 84.0 & 86.2 \\
        
        DAM-VP & \textcolor{gray}{\textit{CVPR 2023}} & ViT-B & 85M & 21K & 88.1 & 97.3 & 73.1 & 87.5 & 92.3 & 90.6 & 99.2 & 87.5 & 82.1 & 86.9 & 88.5 \\

        SA\textsuperscript{2}VP & \textcolor{gray}{\textit{AAAI 2024}} & ViT-B & 85M & 21K & 91.3 & 98.6 & 75.6 & 89.0 & 92.2 & 96.3 & 99.2 & 96.4 & 86.0 & 90.1 & 91.5 \\
        
        AutoVP & \textcolor{gray}{\textit{ICLR 2024}} & CLIP & 85M & 400M & 77.9 & 95.2 & 62.5 & 85.4 & 90.3 & 93.1 & 90.4 & 92.9 & 83.5 & 82.3 & 85.4 \\

        SPT & \textcolor{gray}{\textit{ICML 2024}} & ViT-B & 85M & 21K & 79.2 & 97.6 & 66.5 & 90.6 & 89.8 & 91.3 & 98.3 & 92.1 & 87.6 & 84.3 & 87.7 \\
        
        \specialrule{0.8pt}{1pt}{1pt}
        \specialrule{0.5pt}{1pt}{1pt}
        
        VPT & \textcolor{gray}{\textit{ECCV 2022}} & ViT-S & 22M & 1K & 82.0 & 96.8 & 64.6 & 72.5 & 88.6 & 94.0 & 86.0 & 94.7 & 64.7 &79.6 & 82.4 \\
        
        DAM-VP & \textcolor{gray}{\textit{CVPR 2023}} & ViT-S & 22M & 1K & 86.5 & 97.4 & 69.0 & 78.3 & 86.2 & \underline{96.9} & 86.9 & 96.8 & \underline{75.0} & \underline{85.1} & 85.8\\
        
        AutoVP & \textcolor{gray}{\textit{ICLR 2024}} & ViT-S & 22M & 1K & 72.6 & 92.9 & 56.7 & 49.9 & 78.5 & 87.2 & 67.8 & 89.7 & 38.3 & 60.6 & 69.4 \\
        
        SPT & \textcolor{gray}{\textit{ICML 2024}} & ViT-S & 22M & 1K & 84.2 & 97.1 & \underline{70.6} & 79.1 & 91.8 & 96.0 & \underline{89.1} & 94.5 & 72.7 & 81.0 & 85.6 \\
        
        \hline
        
        Linear & \textcolor{gray}{\textit{-}} & Vim-S & 25M & 1K & 78.1 & 93.8 & 64.5 & 68.1 & 94.8 & 67.7 & 87.0 & 53.4 & 55.0 & 68.0 & 73.0 \\
        
        VPT & \textcolor{gray}{\textit{ECCV 2022}} & Vim-S & 25M & 1K & 84.9 & 96.6 & 69.2 & 77.3 & 94.9 & 92.2 & 88.2 & 93.5 & 69.2 & 79.2 & 84.5 \\
        
        DAM-VP & \textcolor{gray}{\textit{CVPR 2023}} & Vim-S & 25M & 1K & \underline{87.8} & \underline{98.0} & 67.4 & \underline{79.4} & 89.2 & 96.5 & 87.9 & \underline{96.9} & 74.9 & 79.3 & \underline{85.7}\\
        
        AutoVP & \textcolor{gray}{\textit{ICLR 2024}} & Vim-S & 25M & 1K & 76.3 & 95.0 & 56.4 & 54.3 & 87.0 & 82.2 & 73.1 & 82.8 & 52.5 & 62.1 & 72.2\\
        
        SPT & \textcolor{gray}{\textit{ICML 2024}} & Vim-S & 25M & 1K & 84.2 & 96.7 & 68.5 & 74.8 & \textbf{95.0} & 95.8 & 81.6 & 91.1 & 66.9 & 77.8 & 83.2 \\

        \hline
        \rowcolor{gray!20}
        \textbf{SVP(ours)} & \textcolor{gray}{\textit{This Paper}} &Vim-S & 25M & 1K & \textbf{89.8} & \textbf{98.6} & \textbf{74.0} & \textbf{82.8} & \textbf{95.0} & \textbf{97.5} & \textbf{95.2} & \textbf{97.9} & \textbf{78.5} & \textbf{88.2} & \textbf{89.8} \\
    \bottomrule
    \end{tabularx}
    \caption{The comparison results against state-of-the-art methods on HTA benchmark. The best results are bolded, and second-best underlined. 1K and 21K refer to ImageNet-1K and ImageNet-21K, respectively, while 400M refers to the 400 million image-text pairs used in CLIP pre-training.}
\label{tab:hta}
\end{table*}

    From Equations~\ref{EQ11} and~\ref{EQ12}, our method directly activates the update gate~($\mathbf{B}_i(\Delta_i\odot\boldsymbol{x}_i))$ and forget gate~$(\mathbf{\widetilde{A}}_i)$ in Vim, promoting the updation of discriminative information into the hidden state and its propagation across the sequence. This enhances the model's adaptability to new tasks, improving overall performance. Additionally, unlike full fine-tuning, our approach keeps the pre-trained parameters of~$\mathcal{S}_{B}$,~$\mathcal{S}_{C}$, and~$\mathcal{S}_{\Delta}$ fixed. This strategy leverages pre-trained knowledge effectively and mitigates catastrophic forgetting of pre-trained knowledge in downstream tasks.

\section{Experiments}
\subsection{Experiment Setup}
\subsubsection{Datasets and Baselines.} Following prior works~\cite{huang2023diversity, pei2024sa2vp}, our experiments are carried out on two image classification benchmarks HTA and VTAB. 

    \textbf{HTA}. The head tuning adaptation benchmark~\cite{huang2023diversity} comprises 10 datasets including CIFAR10 \cite{krizhevsky2009learning}, CIFAR100~\cite{krizhevsky2009learning}, DTD~\cite{cimpoi2014describing}, CUB-200~\cite{wah2011caltech}, NABirds~\cite{van2015building}, Stanford-Dogs~\cite{khosla2011novel}, Oxford-Flowers~\cite{nilsback2008automated}, Food101~\cite{bossard2014food}, GTSRB~\cite{stallkamp2012man} and SVHN~\cite{netzer2011reading}.

    \textbf{VTAB-1K}. It collects 19 benchmarks from Visual Task Adaptation~\cite{zhai2019large}, categorized into three groups: i) Natural, ii) Specialized, and iii) Structured, each with 1000 training examples. Following~\cite{zhai2019large, jia2022visual}, we use an 800-200 train/val split.

\begin{table*}[!h]
\small
    \centering
    \setlength{\tabcolsep}{0.57mm}
    \renewcommand{\arraystretch}{1.3}
    \begin{tabular}{c|c|ccccccc|c|cccc|c|cccccccc|c|c}
        \toprule
         \multirow{5}{*}{\rotatebox{90}{Methods}} & \multirow{5}{*}{\rotatebox{90}{Backbone}} & \multicolumn{8}{c|}{Natural} & \multicolumn{5}{c|}{Specialized} & \multicolumn{9}{c|}{Structured} & \multirow{5}{*}{\textbf{\rotatebox{90}{Average}}}\\
        \cline{3-24}
          & & \rotatebox{90}{\fontsize{8pt}{12pt}\selectfont CIFAR100} & \rotatebox{90}{\fontsize{8pt}{12pt}\selectfont Caltech101} & \rotatebox{90}{\fontsize{8pt}{12pt}\selectfont DTD} & \rotatebox{90}{\fontsize{8pt}{12pt}\selectfont Flowers102} & \rotatebox{90}{\fontsize{8pt}{12pt}\selectfont Pets} & \rotatebox{90}{\fontsize{8pt}{12pt}\selectfont SVHN} & \rotatebox{90}{\fontsize{8pt}{12pt}\selectfont Sun397} & \rotatebox{90}{\fontsize{8pt}{12pt}\selectfont Average}& \rotatebox{90}{\fontsize{8pt}{12pt}\selectfont Camelyon} & \rotatebox{90}{\fontsize{8pt}{12pt}\selectfont EuroSAT} & \rotatebox{90}{\fontsize{8pt}{12pt}\selectfont Resisc45} & \rotatebox{90}{\fontsize{8pt}{12pt}\selectfont Retinopathy} & \rotatebox{90}{\fontsize{8pt}{12pt}\selectfont Average} & \rotatebox{90}{\fontsize{8pt}{12pt}\selectfont Clevr-Count} & \rotatebox{90}{\fontsize{8pt}{12pt}\selectfont Clevr-Dist} & \rotatebox{90}{\fontsize{8pt}{12pt}\selectfont DMLab} & \rotatebox{90}{\fontsize{8pt}{12pt}\selectfont KITTI-Dist} & \rotatebox{90}{\fontsize{8pt}{12pt}\selectfont dSpr-Loc} & \rotatebox{90}{\fontsize{8pt}{12pt}\selectfont dSpr-Ori} & \rotatebox{90}{\fontsize{8pt}{12pt}\selectfont sNORB-Azim} & \rotatebox{90}{\fontsize{8pt}{12pt}\selectfont sNORB-Ele} & \rotatebox{90}{\fontsize{8pt}{12pt}\selectfont Average} & \\
        \hline
        Full & Vim-S & 49.9 & 86.9 & 66.5 & 93.2 & 91.6 & 88.3 & 40.3 & 73.8 & 86.3 & 93.9 & 83.7 & 76.4 & 85.1 & 54.3 & 48.3 & 50.5 & 52.7 & 36.0 & 34.9 & 28.5 & 32.9 & 42.3 & 67.1 \\
        \hline
        Linear & Vim-S & 47.0 & 84.2 & 60.7 & 77.1 & 90.0 & 42.3 & 39.8 & 63.0 & 78.6 & 87.2 & 71.1 & 73.8 & 77.7 & 32.4 & 33.9 & 35.8 & 50.9 & \underline{49.6} & \underline{49.3} & 20.0 & 22.6 & 36.8 & 59.2 \\
        
        VPT & Vim-S & \underline{57.1} & \underline{86.4} & 65.0 & \underline{86.8} & \underline{90.8} & \underline{78.0} & \underline{42.0} & \underline{72.3} & \underline{80.4} & \underline{90.3} & \underline{78.1} & 74.4 & \underline{80.8} & \underline{36.2} & \underline{40.2} & 36.3 & 46.6 & 44.8 & 29.5 & 20.0 & \underline{29.0} & 35.3 & \underline{62.8} \\

        SPT & Vim-S & 54.0 & 85.5 & \textbf{68.5} & 82.8 & 89.7 & 72.0 & 40.0 & 70.4 & 79.8 & 89.8 & 75.2 & \underline{74.5} & 79.8 & 35.0 & 38.0 & \underline{38.2} & \underline{51.2} & 38.8 & 40.1 & \underline{20.9} & 24.1 & \underline{35.8} & 62.0 \\
        
        \hline
        \rowcolor{gray!20}
        
        \textbf{SVP(ours)} & Vim-S & \textbf{59.3} & \textbf{87.4} & \underline{67.0} & \textbf{92.4} & \textbf{92.8} & \textbf{88.5} & \textbf{43.2} & \textbf{75.8} & \textbf{86.0} & \textbf{94.7} & \textbf{84.2} & \textbf{77.1} & \textbf{85.5} & \textbf{54.1} & \textbf{52.0} & \textbf{52.0} & \textbf{51.8} & \textbf{62.3} & \textbf{54.2} & \textbf{30.7} & \textbf{35.4} & \textbf{49.1} & \textbf{70.1} \\
        \bottomrule
    \end{tabular}
    \caption{The comparison results against state-of-the-art methods on VTAB-1K benchmark. The best results are bolded, and second-best underlined. Overall ``Average” is the group-wise average accuracy over three groups.}
    \label{tab:VTAB}
\end{table*}

    \textbf{Comparison Methods.} We compare our SVP with other visual prompting methods including VPT~\cite{jia2022visual}, DAM-VP~\cite{huang2023diversity}, AutoVP~{\cite{tsao2024autovp}} and SPT~\cite{wang2024revisiting}. We apply these methods to Vision Mamba~\cite{zhu2024vision}. Additionally, We compare the performance of these prompting methods using ViT-Small~\cite{dosovitskiy2020image} as the backbone, with the same model size and pre-trained dataset to Vision Mamba.
    We also present the visual prompting results in ViT-B with much larger parameters for reference.

    \textbf{Implementation Details.}~Our experiments primarily involve three pre-trained vision models: ViT-Small/16 and Vim-Small, both of which are pre-trained on ImageNet-1K~\cite{russakovsky2015imagenet}, and ViT-Base/16~\cite{dosovitskiy2020image}, which is pre-trained on ImageNet-21K~\cite{krizhevsky2012imagenet}. Following~\cite{huang2023diversity}, all methods are trained for 100 epochs across all datasets for a fair comparison. For the compared methods, we use the optimizers specified in the original papers to achieve better performance. In our approach, we utilize the AdamW~\cite{loshchilov2017decoupled} optimizer for optimization and implement cosine annealing.
    The number of shared layers in Cross-Prompting is set to 4, 8, or 12, depending on the dataset, and the hidden dimension of the inner-prompts generator is set to 64.

\setlength{\tabcolsep}{1.3mm}\setlength{\extrarowheight}{1.25pt}
\begin{table}[tb]
\small
    \centering
    \begin{tabular}{c|c|c|c|c|c}
    \toprule
    Format & Position & Cifar & DTD & CUB & Flowers \\
    \hline
    \multirow{5}{*}{Append} & Pre         & 84.9 & 69.2 & 77.3 & 88.4 \\
                            & Post        & 84.9 & 69.9 & 77.2 & 88.0 \\
                            & Both Sides  & 85.0 & 69.7 & 77.4 & 89.3 \\
                            & Uniform     & 84.4 & 69.5 & 77.0 & 88.0 \\
                            & Middle      & 85.3 & 69.7 & 77.6 & 88.9 \\
    \hline
    \rowcolor{gray!20}
    \multicolumn{2}{c|}{\textbf{SVP(Ours)}}& \textbf{89.8} & \textbf{74.0} & \textbf{82.8} & \textbf{95.2}\\
    \bottomrule
    \end{tabular}
    \caption{Ablation of prompt formation and position. Append indicates a prompt sequence inserted into the token sequence, with Position specifying the insertion location.}
    \label{tab:table3}
\end{table}

\setlength{\extrarowheight}{2pt}
\begin{table}[tb]
\small
    \centering
    \begin{tabular}{>{\centering\arraybackslash}p{0.9cm}|>{\centering\arraybackslash}p{0.9cm}|c|c|c|c}
    \toprule
        \multicolumn{2}{c|}{Component} & & & & \\
        \cline{1-2}
        
        $\mathbf{IP}$ & 
        $\mathbf{CP}$ & 
        \multirow{-2}{*}{Cifar} & \multirow{-2}{*}{DTD} & \multirow{-2}{*}{CUB} & \multirow{-2}{*}{Flowers} \\
    \hline
        - & -                   & 78.1 & 64.5 & 68.1 & 86.0 \\
        - & \ding{51}           & 87.2 & 72.5 & 80.0 & 91.8\\
        \ding{51} & -           & 89.6 & 71.6 & 82.1 & 93.2\\
        \hline
        \rowcolor{gray!20}
        \ding{51} & \ding{51}  & \textbf{89.8} & \textbf{74.0} & \textbf{82.8} & \textbf{95.2}\\
    \bottomrule
    \end{tabular}
    \caption{Ablation results of Cross-Prompting~($\mathbf{CP}$) and Inner-Prompting~($\mathbf{IP}$).  - and \ding{51} represent without or with the~component.}
    \label{tab:components}
\end{table}

\subsection{Comparison with State-of-the-arts}

    \textbf{Results on HTA.} We first conduct experiments on HTA datasets using the ImageNet-1k supervised ViT-Small/16 and Vim-Small as the pre-trained models. 
    As shown in Table~\ref{tab:hta}, our SVP significantly surpasses the results of existing prompting methods that use pre-trained models with the same model size and pre-training dataset. It not only achieves SOTA performance in average accuracy but also excels across nine out of ten datasets. 
    For example, SVP achieves a notable improvement of \textbf{5.3\%} over VPT when applied to Vim, while also surpassing DAM-VP by \textbf{4.1\%}. This is because our SVP effectively activates the update and forget gates of Vim across the whole sequence, promoting the propagation of discriminative information, which leads to enhanced performance.
    
    Additionally, compared to full fine-tuning, our method outperforms in 7 out of 10 datasets and exceeds 1.3\% in average. This further supports the discussion in our methodology section, demonstrating that our approach mitigates the issue of catastrophic forgetting commonly seen in full fine-tuning. Our SVP retains more pre-trained knowledge while efficiently adapting to downstream tasks.

    Notably, our method achieves performance comparable to prompting methods that use ViT-B as the backbone. ViT-B has a much larger model size of 85 Million parameters and is pre-trained on the much larger dataset ImageNet-21K~\cite{deng2009imagenet}. In contrast, our method uses a significantly smaller model and a smaller pre-training dataset, yet still delivers comparable results. Our approach outperforms methods such as DAM-VP~\cite{huang2023diversity} and SPT~\cite{wang2024revisiting}. This can be attributed to the dual-path Selective Prompting design in Vim, which effectively promotes the propagation of both shared inter-layer information and specific intra-layer information.

\begin{figure}[t]
    \centering
    \includegraphics[width=0.48\textwidth]{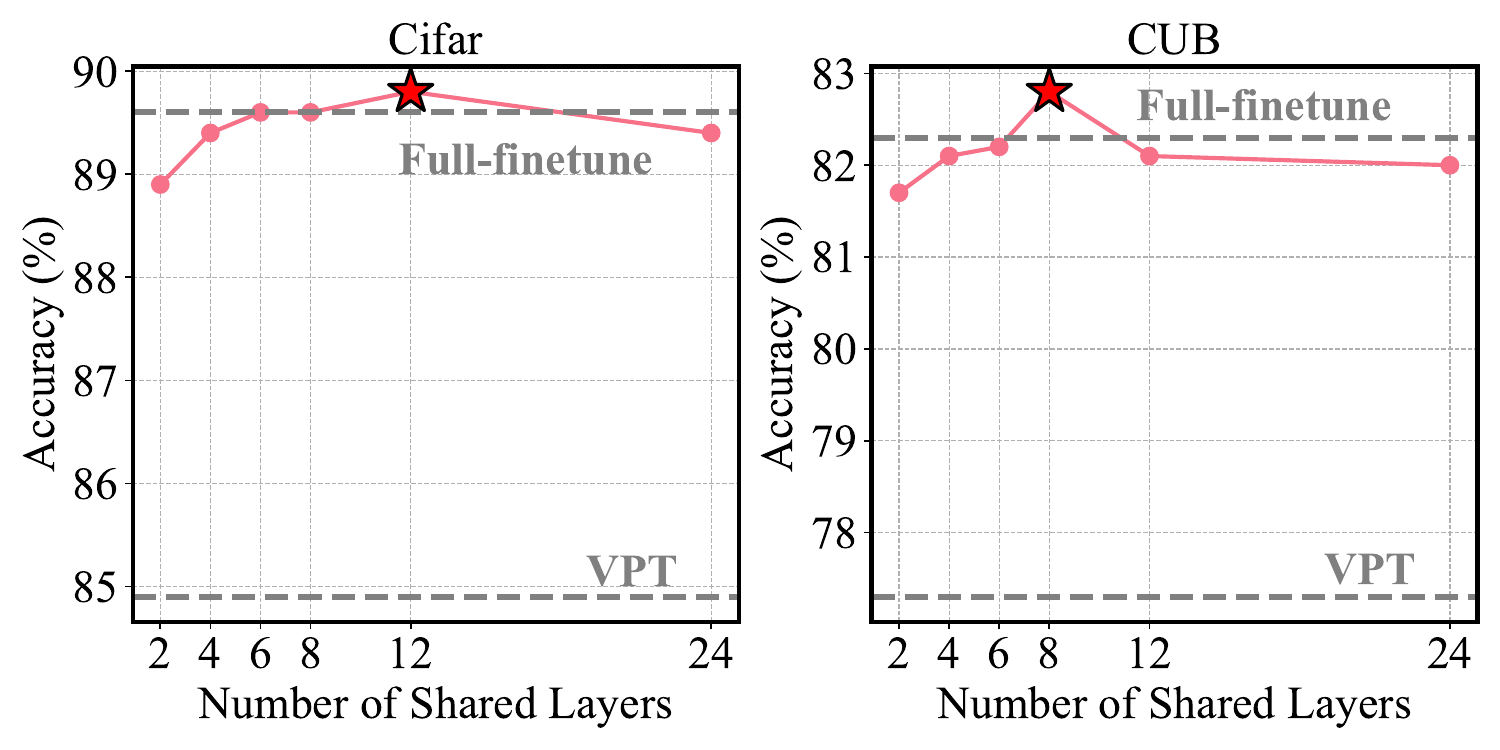} 
    \caption{Ablation results of the number of shared layers in Cross-Prompting.}
    \label{fig4}
\end{figure}

    \textbf{Results on VTAB-1K.} Following~\cite{jia2022visual, pei2024sa2vp}, we also conduct experiments on another widely used VTAB-1k~\cite{zhai2019large} benchmark. As shown in Table~\ref{tab:VTAB}, compared to the method VPT~\cite{jia2022visual}, our SVP achieves improvements of \textbf{3.5\%}, \textbf{4.7\%}, and \textbf{13.8\%} in the three different tasks Natural, Specialized, and Structured, respectively. This further illustrates the effectiveness of our SVP in promoting the information flow and the robust adaptability of our SVP.

\subsection{Ablation Study}

     \textbf{Ablation of Prompt Format and Position.} Given the specificity of Vim's linear sequence model, where token impact varies by position, we first explore appending prompts to the image sequence and assess the effect of changing their position. As shown in Table~\ref{tab:table3}, placing the prompt tokens in the middle yields a slight average improvement of 0.5\% over in the pre, likely due to its closeness to the class token. However, this approach does not fully consider the sequencial token-wise compression and propagation characteristics of the Vim sequence model. This limitation makes it ineffective in learning the input distribution across the entire sequence and in activating Vim's update and forget gates. In contrast, our SVP attains an average 
     \textbf{5.2}\% improvement. It is because our SVP generates token-wise prompts based on the input, more effectively learning the input distribution. This selective change activates the update and forget gates in Vim, promoting the propagation of discriminative information.

\begin{figure}[t]
    \begin{center}
    \includegraphics[width=0.47\textwidth]{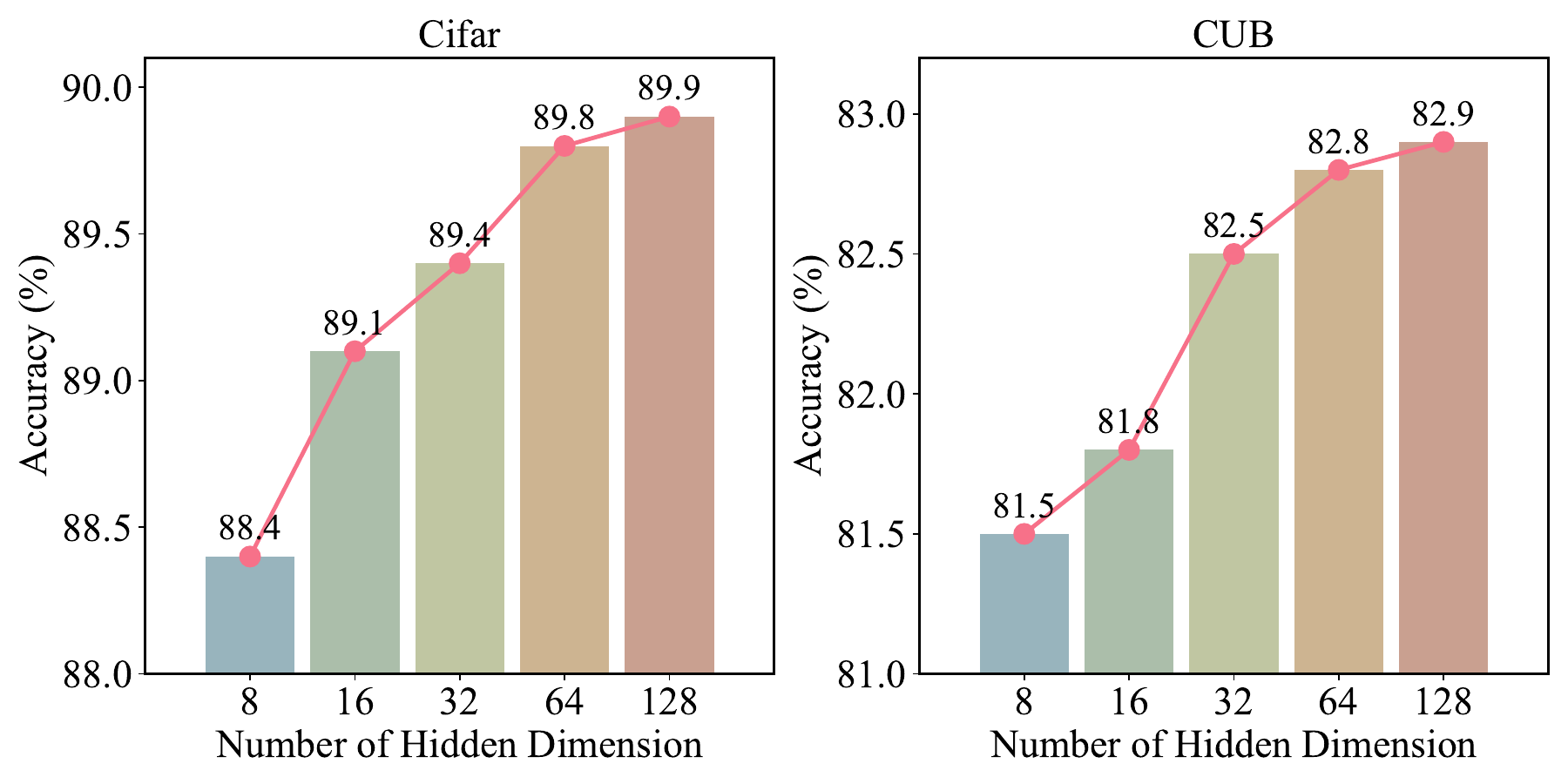}
    \end{center}
    \caption{Ablation of hidden dimension in Inner-Prompting.}
    \label{fig5}
\end{figure}

\begin{figure}[t]
    \begin{center}
    \includegraphics[width=0.47\textwidth]{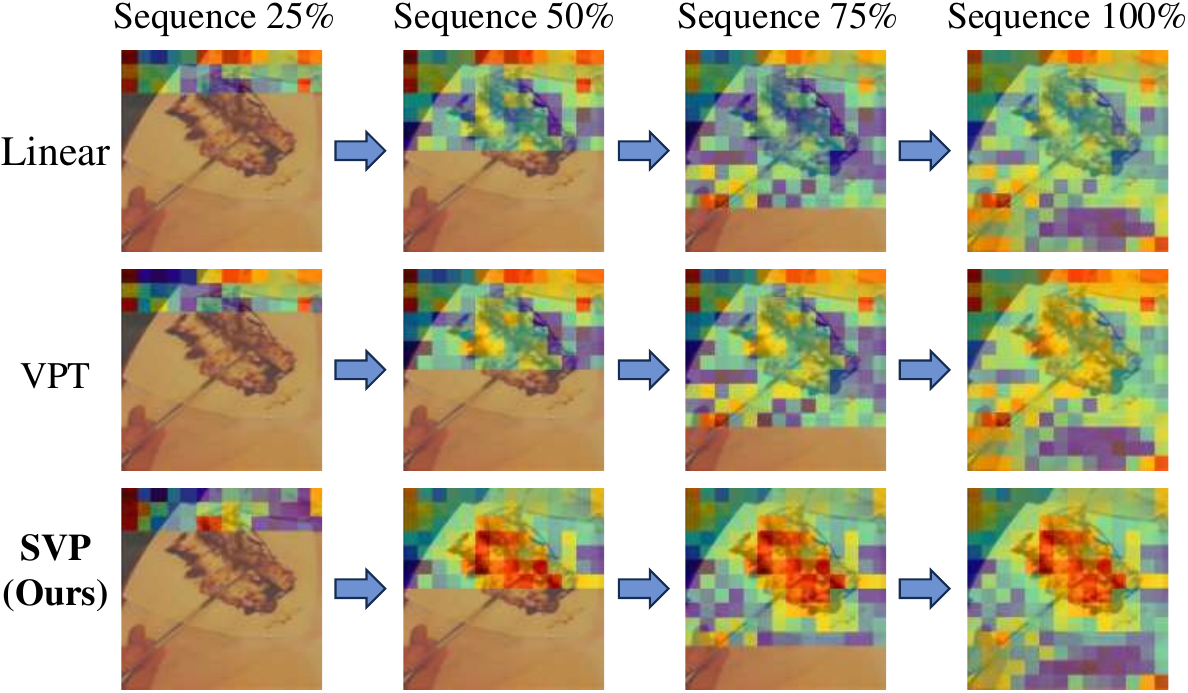}
    \end{center}
    \caption{Visualization of normalized update gate of Vim over 24 layers during sequence propagation. It reveals the information retained in Vim's hidden states as the sequence propagates to 25\%, 50\%, 75\%, and the end of sequence.}
    \label{fig6}
\end{figure}

    \textbf{Influence of Different Components.} 
    To evaluate the effectiveness of dual-path prompting in our proposed SVP, we conduct ablation experiments on four datasets, as shown in Table~\ref{tab:components}. When no prompts are used, SVP reduces to a frozen pre-trained Vim model with a learnable classifier. On the CUB dataset, employing only the inner-prompts~$\boldsymbol{p}^{\scriptscriptstyle I}$ boosts performance by 14\%. Furthermore, using both the inner-prompts~$\boldsymbol{p}^{\scriptscriptstyle I}$ and cross-prompts~$\boldsymbol{p}^{\scriptscriptstyle C}$ together yields an additional 0.7\% improvement. 

    This is because our dual-path SVP method captures more discriminative and richer information compared to the single-path approach. This enhancement arises from the synergy between the shared information provided by Cross-Prompting and the layer-specific details from Inner-Prompting, enabling more precise extraction and propagation of discriminative information. 

    \textbf{Influence of Hyper-Parameters.} 
    The shared layers in Cross-Prompting and the hidden dimension of the inner-prompts generator ($\mathcal{G}^{\scriptscriptstyle I}$) are important hyper-parameters in our SVP. To assess their impact, we conduct extensive ablation experiments. As shown in Figure \ref{fig4}, the model’s performance initially improves but then declines as the number of shared layers increases, corresponding to the shallow, intermediate, and deep layers, which tend to share information respectively. For hidden dimension experiments in Figure \ref{fig5}, performance continues to improve with increasing rank, but the rate of improvement slows. To balance performance and tunable parameters, we select 64, resulting in 1.6M tunable parameters. For those willing to trade a small amount of performance, a dimension of 32, with 0.9M tunable parameters, is also an option.

\begin{figure}[tbp]
    \begin{center}
    \includegraphics[width=1.0\linewidth]{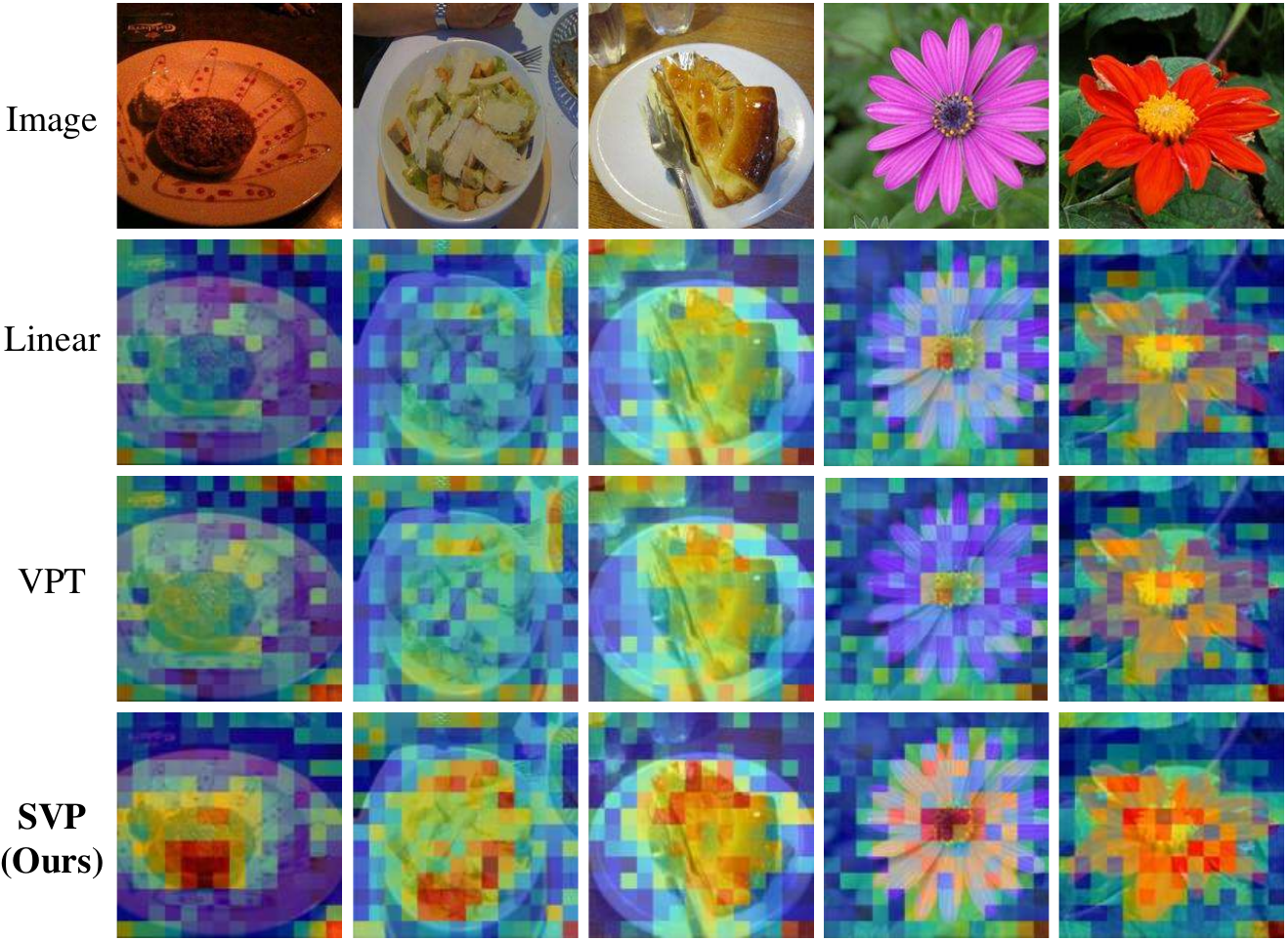}
    \end{center}
    \caption{Visualization of the average values of the update gate over 24 layers in Vim.}
    \label{fig7}
\end{figure}
    
    \textbf{Visualization Results of Update Gate in Vim.} As shown in Figure~\ref{fig6}, we first visualize the normalized update gate of Vim during the sequence propagation. 
    This visualization reveals the information retained in Vim's hidden states as the sequence progresses, highlighting that our method improves the extraction and transmission of discriminative features. 
    Figure~\ref{fig7} shows more examples of the update gate within Vim. Our SVP retains more discriminative information related to key elements like flowers and food, while less discriminative backgrounds, such as plates, are retained to a lesser extent. The visualization results demonstrate that our method effectively activates the update gate in downstream tasks, enabling Vim to focus more on discriminative regions and thereby enhancing feature extraction and propagation.

\section{Conclusion}

    In this paper, we introduce Selective Visual Prompting (SVP), an efficient and novel visual prompting method tailored for Vim. To the best of our knowledge, this is the initial exploration of visual prompting within Vim. Unlike existing approaches, our SVP leverages a dual-path strategy to achieve superior performance by leveraging both shared and layer-specific information. We find that prompts selectively generated based on input are more effective in activating Vim's update and forget gates, promoting information propagation. Visualization results further validate our approach. We believe SVP will serve as a valuable benchmark that will drive future research in visual prompting for Vim.

\section{Acknowledgments}
This work was supported by the grants from the National Natural Science Foundation of China (62376011, 61925201, 62132001, 62432001) and Beijing Natural Science Foundation (L247006).
\bigskip

\bibliography{aaai25}

\end{document}